\documentclass[aip,cha,reprint]{revtex4-1}
\usepackage{graphicx}
\usepackage{color}
\usepackage{amsmath}
\usepackage{graphicx}
\usepackage{float}
\usepackage{subfigure}

\begin{document}
\title{Performing edge detection by difference of Gaussians using q-Gaussian kernels}

\author{Lucas Assirati} 
\affiliation{Scientific Computing Group, S\~ao Carlos Institute of Physics, University of S\~{a}o Paulo (USP),  cx 369 13560-970 S\~{a}o Carlos, S\~{a}o Paulo, Brazil - www.scg.ifsc.usp.br}
\email{assirati@usp.br, bruno@ifsc.usp.br}
\author{N\'ubia R. da Silva} 
\affiliation{Scientific Computing Group, S\~ao Carlos Institute of Physics, University of S\~{a}o Paulo (USP),  cx 369 13560-970 S\~{a}o Carlos, S\~{a}o Paulo, Brazil - www.scg.ifsc.usp.br}
\affiliation{Institute of Mathematics and Computer Science, University of S\~{a}o Paulo (USP), Avenida Trabalhador s\~{a}o-carlense, 400 13566-590 S\~{a}o Carlos, S\~{a}o Paulo, Brazil}
\email{[nubiars, lberton, alneu]@icmc.usp.br}
\author{Lilian Berton}  
\affiliation{Institute of Mathematics and Computer Science, University of S\~{a}o Paulo (USP), Avenida Trabalhador s\~{a}o-carlense, 400 13566-590 S\~{a}o Carlos, S\~{a}o Paulo, Brazil}
\author{Alneu de A. Lopes} 
\affiliation{Institute of Mathematics and Computer Science, University of S\~{a}o Paulo (USP), Avenida Trabalhador s\~{a}o-carlense, 400 13566-590 S\~{a}o Carlos, S\~{a}o Paulo, Brazil}
\author{Odemir M. Bruno}
\affiliation{Scientific Computing Group, S\~ao Carlos Institute of Physics, University of S\~{a}o Paulo (USP),  cx 369 13560-970 S\~{a}o Carlos, S\~{a}o Paulo, Brazil - www.scg.ifsc.usp.br}
\affiliation{Institute of Mathematics and Computer Science, University of S\~{a}o Paulo (USP), Avenida Trabalhador s\~{a}o-carlense, 400 13566-590 S\~{a}o Carlos, S\~{a}o Paulo, Brazil}

\begin{abstract}
In image processing, edge detection is a valuable tool to perform the extraction of features from an image. This detection reduces the amount of information to be processed, since the redundant information (considered less relevant) can be unconsidered. 
The technique of edge detection consists of determining the points of a digital image whose intensity changes sharply. This changes are due to the discontinuities of the orientation on a surface for example.  A well known method of edge detection is the Difference of Gaussians (DoG). The method consists of subtracting two Gaussians, where a kernel has a standard deviation smaller than the previous one. The convolution between the subtraction of kernels and the input image results in the edge detection of this image.  This paper introduces a method of extracting edges using DoG with kernels based on the q-Gaussian probability distribution, derived from the q-statistic proposed by Constantino Tsallis. To demonstrate the method's potential, we compare the introduced method with the traditional DoG using Gaussians kernels. The results showed that the proposed method can extract edges with more accurate details.
\end{abstract}

\keywords{edge Detection, difference of Gaussians, q-Gaussian}

\maketitle

\section{Introduction}
Image processing is designated as any type of signal processing where the input is an image and the output can be another image or a set of features extracted from the input image. Once the computer vision involves the identification and subsequent classification of certain objects in a given image, edges detection is an essential tool in image analysis. When performing edge detection on an image, there is a reduction of the amount of information to be processed because the redundant information (considered less relevant) can be unconsidered.

The segmentation by edge detection is based on two important concepts: similarity and discontinuity. Thus, the algorithms look for points (or curves and contours) of the digital image where the intensity changes abruptly. This sudden change in intensity may occur for various reasons, as example, the orientation discontinuities in a surface and changes in brightness and illumination in a scene.
Applications for the edge detection method are found in various fields of science: medicine \cite{Gudmundsson1998}, engineering and satellite images \cite{Augusto1984}, robotics and machine vision\cite{jain1995machine}.

There are several methods for edge detection, like: Canny, Sobel, Prewitt, and based on Gaussian masks (kernels), as Laplacian of Gaussian (LoG) and Difference of Gaussian (DoG) \cite{gonzalez2011digital}. The DoG method generally uses classical Gaussians in its approach. But this work suggests the use of q-Gaussian for the composition of the mask that will be applied to the image to extract its edges. The q-Gaussian probability distribution comes from the q-algebra introduced by Tsallis. 

The q-algebra is derived from Tsallis definition of non-extensive entropy. There are some works in the literature that used with success the Tsallis q-entropy into the image processing \cite{PortesdeAlbuquerque20041059,Sathya2010,Kilic2012} and image analysis \cite{Fabbri2013,Fabbri2012} fields. The q-Gaussian kernels was previously used to noise reduction \cite{Soares2013}. In this work, we present a proposal of compose the DoG filter using the q-Gaussian kernels. The potential of the proposed method is demonstrated by comparing with the traditional method for DoG. One can notice that it is able to perform edge extraction with one more parameter (q), which can make the DoG approach more flexible.  

\begin{figure}[!htbp]
	\center
		\subfigure[]{\label{fig:1}\includegraphics[width= .40\textwidth]{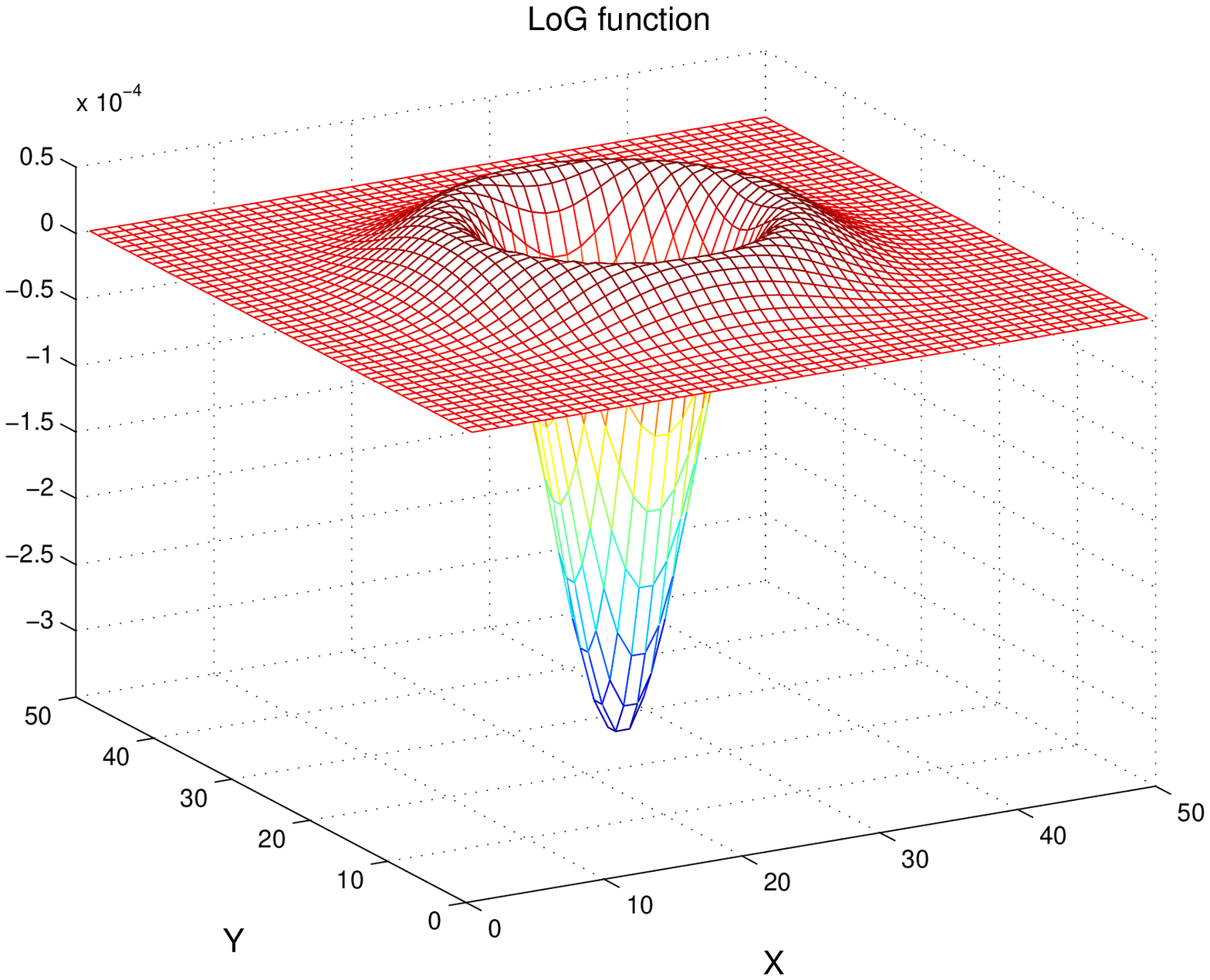}}
		\subfigure[]{\label{fig:2}\includegraphics[width= .40\textwidth]{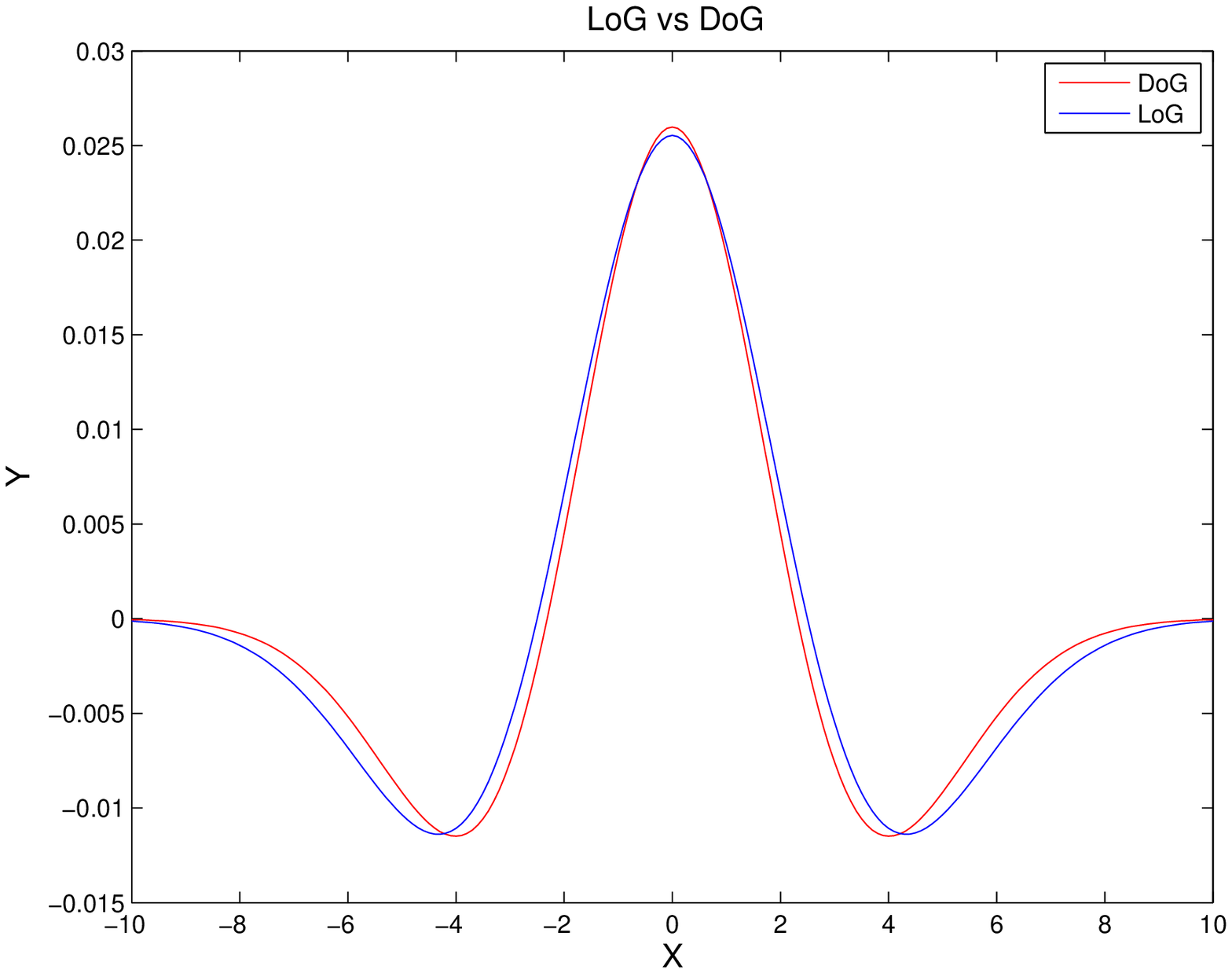}}
	\caption{(a)LoG Function, $\sigma = 2.5$, (b) Difference of Gaussians vs Laplacian of Gaussians 1D.}
	\label{fig:1e2}
\end{figure}

\section{Laplacian of Gaussian vs Difference of Gaussians}
Consider the one-dimensional Gaussian distribution:
$f(x,\mu, \sigma) = \frac{1}{\sqrt{2 \pi \sigma^2}} \exp{\left(-\frac{(x-\mu)^2)}{2 \sigma^2}\right)} ,$ with $-\infty < x < \infty,$ and $\sigma > 0$, 
where $\mu$ is the mean, $\sigma$ is the standard deviation and $\sigma^2$ is the variance.

If we take the second derivative of the one-dimensional Gaussian function considering $\mu = 0$, we obtain the Ricker wavelet:
$\psi(x) = \frac{1}{\sqrt{2 \pi \sigma^2}} \frac{1}{\sigma^4} (x^2 - \sigma^2) \exp{\frac{-x^2}{2\sigma^2}}$.

In the two-dimensional the Gaussian distribution becomes:
\begin{equation}
\begin{split}
f(x, y, \sigma) = \frac{1}{2 \pi \sigma^2} \exp{\left(-\frac{(x^2+y^2)}{2 \sigma^2}\right)} ,\\ where -\infty < x, y < \infty, \sigma > 0
\end{split}
\end{equation}

The Laplacian of Gaussian LoG is a multidimensional generalization of the Ricker wavelet. To obtain it we need to take the two-dimensional Laplacian of the Gaussian distribution:

\begin{equation}
LoG(x,y) = - \frac{1}{\pi\sigma^4}\left[1-\frac{x^2+y^2}{2\sigma^2}\right] \exp{-\frac{x^2 + y^2}{2\sigma^2}}
\end{equation}

However, in practice the Laplacian of Gaussian (LoG), Figure \ref{fig:1}, is approximated by the Difference Gaussians (DoG) function since this reduces the computational costs in two or more dimensions. The DoG is obtained by performing the subtraction of two Gaussian kernels where a kernel must have a standard deviation slightly lower than its previous.


Figure \ref{fig:2} compares the LoG function with $\sigma=2.5$ with the DoG function using kernels with $\sigma_1 = 2.5$ and $\sigma_2 = 2.15$. The DoG method has lower computational cost, what justifies its use in this study. The convolution of the DOG filter with the input image generates the edge detection for this image.
 
\section{q-Gaussian}

\begin{figure}[!htbp]
	\center
	\includegraphics[width=8cm]{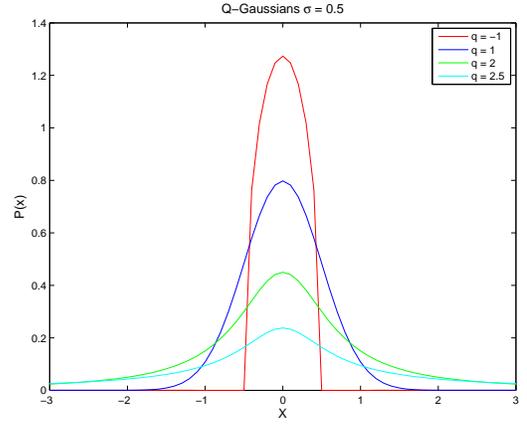}
	\caption{Q-Gaussian Function with $\sigma = 0.5$.}
	\label{fig:3}
\end{figure}

In 1988, Tsallis proposed the non additive statistical mechanics, entitled ``Q-statistic''\cite{Tsallis:1988ws}. This theory suggests that different systems require different tools of analysis, appropriated to the particularities of this system. The informational tool entropy, applied to the information theory by Shannon\cite{Shannon:1948wk} is defined as: $S(x) = - \sum_{x =0}^{W} p(x)\log p(x)$, where $p(x)$ is the occurrence probability, and $W$ is the total number of probabilities.

The generalization proposed by Tsallis gives the definition of the q-entropy: $S_q (x) = \frac{1}{q-1} \left(1 - \sum_{x =0}^{W} p^q (x) \right)$,
where $p(x)$ is the occurrence probability, $W$ is the total number of probabilities and $q$ is an adjustable parameter, freely variable. The correct choice of certain $q$ parameters can evidence important characteristics of the system. When $q \rightarrow 1$, one retrieves the standard entropy.

The q-Gaussian probability distribution comes from the maximization of the Tsallis entropy under appropriate constraints\cite{Tsallis:2011hr}. Again, when $q \rightarrow 1$, one retrieves the Gaussian distribution. The q-Gaussian is defined as:

\begin{equation}
G_q (x) = \frac{1}{C_q \sqrt{2 \sigma^2}} \exp_q{\frac{-x^2}{2\sigma^2}}
\end{equation}

with $\exp_q(x) = [1+(1-q)x]^\frac{1}{1-q}$ and
\[
    C_q = 
\begin{cases}
    \frac{2\sqrt{\pi} \Gamma (\frac{1}{1-q})}{(3-q)\sqrt{1-q}\Gamma(\frac{3-q}{2(1-q)})} & \text{if } -\infty < q < 1\\
    \sqrt{\pi} & \text{if } q = 1\\
    \frac{\sqrt{\pi} \Gamma (\frac{3-q}{2(q-1)})}{(3-q)\sqrt{q-1}\Gamma(\frac{1}{q-1})} & \text{if } -\infty < q < 1
\end{cases}
\]

Figure \ref{fig:3} shows some of the curves generated by the equations of q-Gaussian, compared with the classical Gaussian ($q = 1$):

It is important to note that all the curves have the same parameter $\sigma = 0.5$, but through the generalization proposed by Tsallis, we gain a second adjustable parameter, $q$. Changes in this parameter are able to promote changes in the traditional Gaussian shape, adapting it to the peculiarities of the problem in which it is applied. When $q \rightarrow -\infty$, the one-dimensional q-Gaussian function tends to the Dirac function. Moreover the shape of the q-Gaussian function tends to a straight line $P(x) = 0$ when $q$ approaches the value $3$.

\begin{figure}[!htpb]
	\center
		\subfigure[ $q=1$, $\sigma = 1$]{\label{fig:4}\includegraphics[width= .32\textwidth]{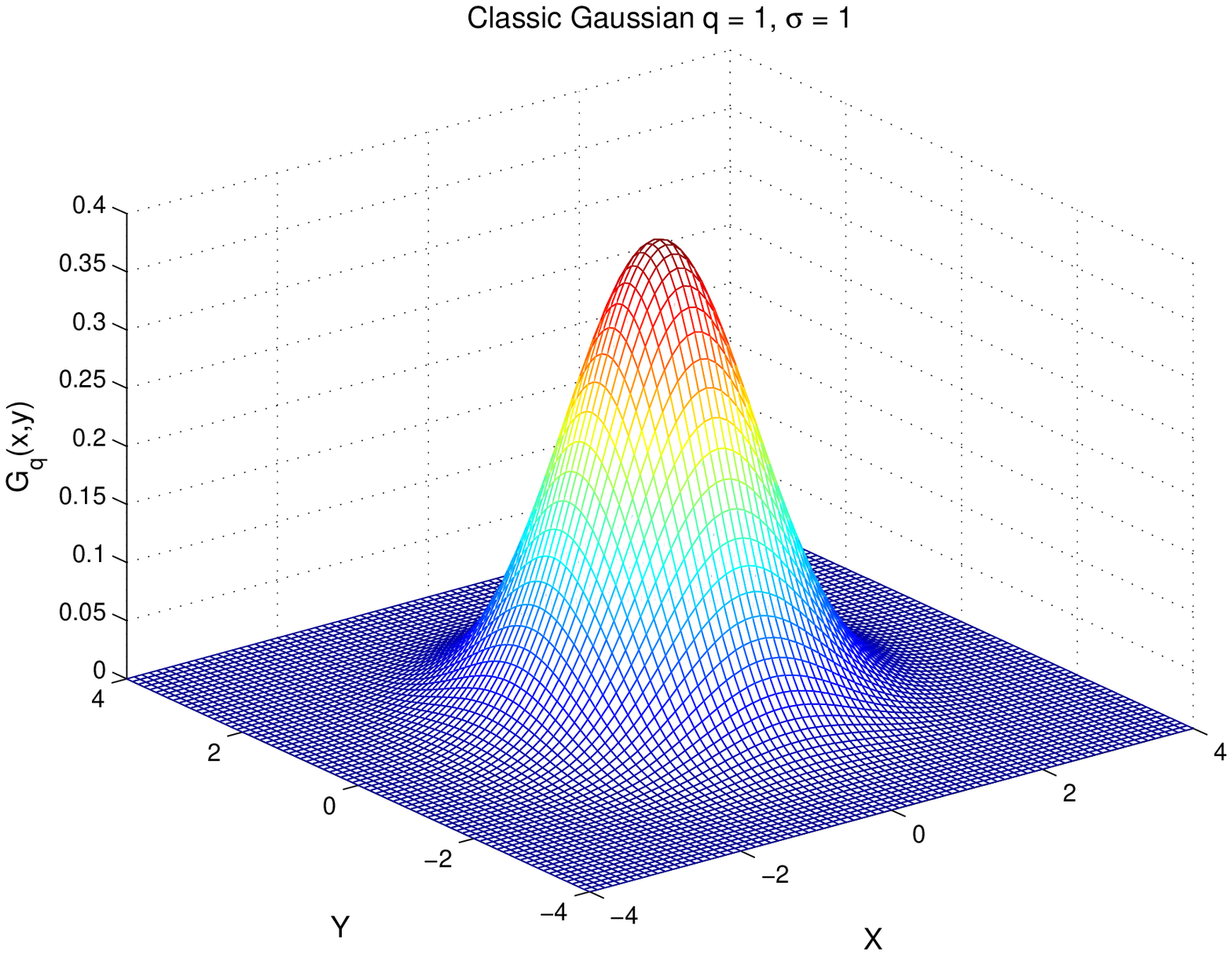}}
		\subfigure[$q=-1$, $\sigma = 1$]{\label{fig:5}\includegraphics[width= .32\textwidth]{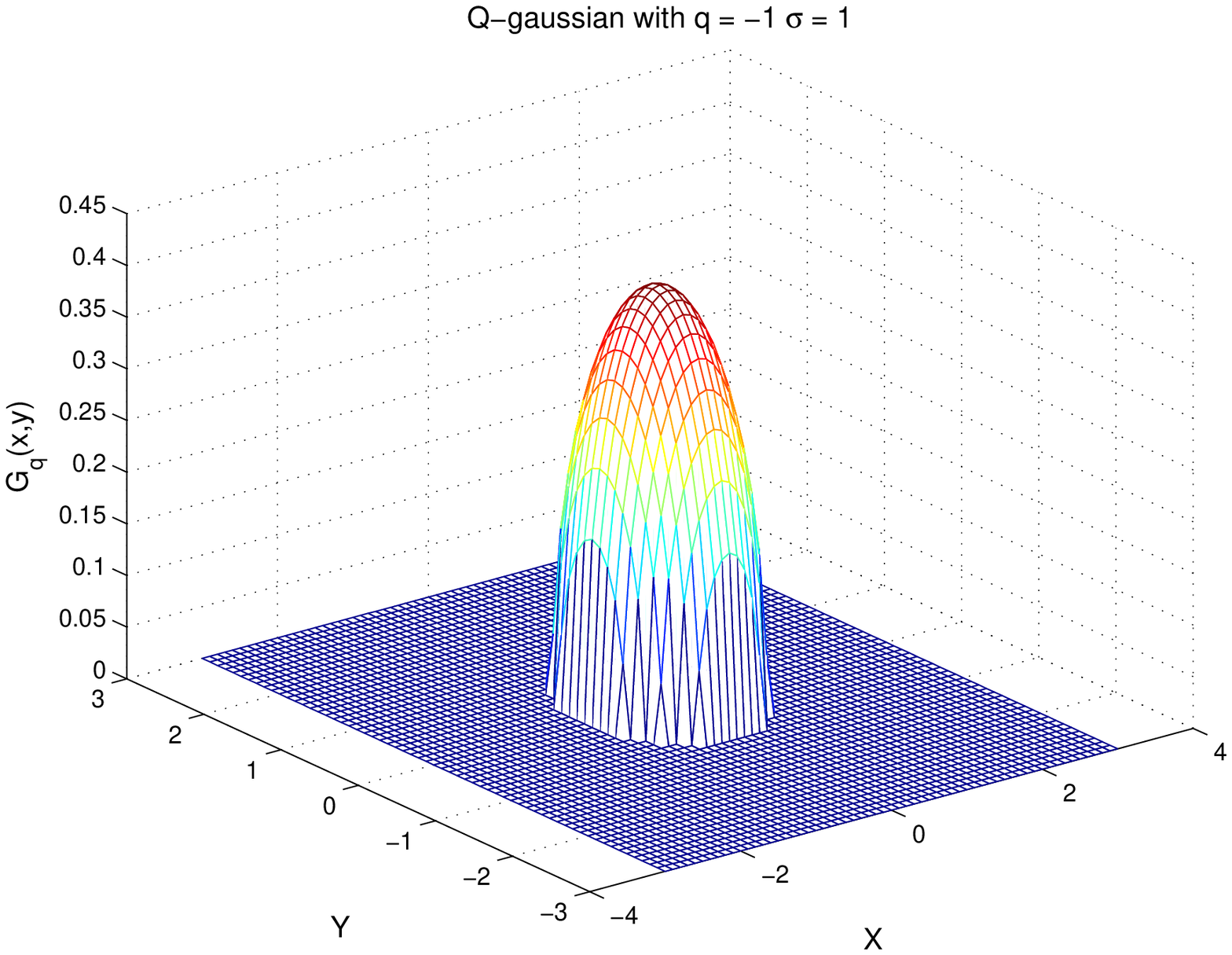}}
		\subfigure[$q=2.5$, $\sigma = 1$]{\label{fig:6}\includegraphics[width= .32\textwidth]{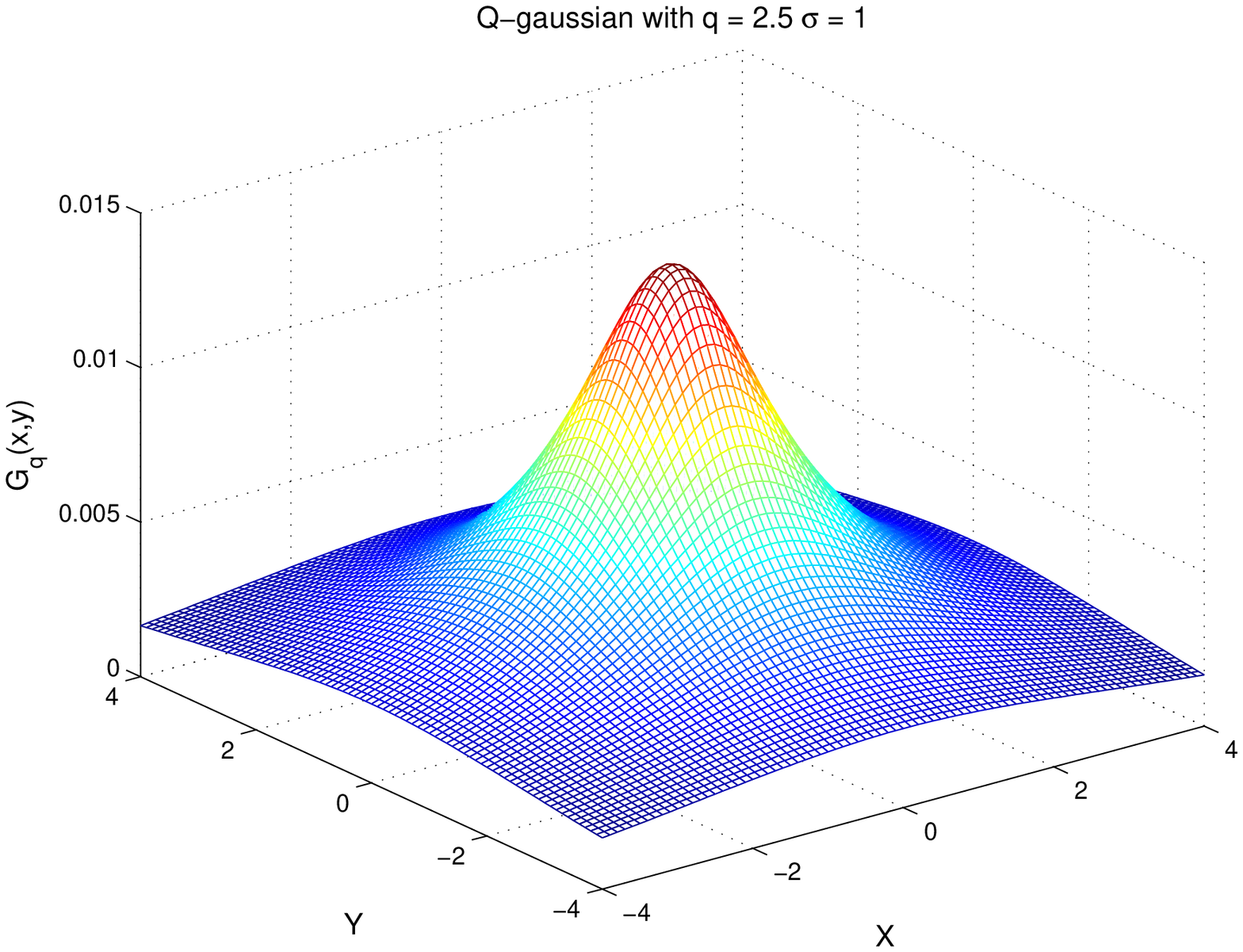}}
	\caption{Q-Gaussian Function.}
	\label{fig:gaussian-function}
\end{figure}

The same way as the classical Gaussian has a Two-Dimensional version, we can derive the multidimensional generalization to q-Gaussian. The bi-dimensional q-Gaussian 
is defined by $G_q (x,y) = \frac{\exp_q{-(x^2+y^2)/(2\sigma^2)}}{2 C_q^2 \sigma^2} $. It is important to note that curves with the same parameter $\sigma$ can have its shape changed adjusting the $q$ parameter, adapting it to the peculiarities of the problem in which it is employed. Figures \ref{fig:4}, \ref{fig:5} and \ref{fig:6} show some representatives of the family of 2D q-Gaussian.

%
%
%

\section{Method}
This work introduces the use of DoG method using q-Gaussian kernels as an alternative to traditional use of Gaussian kernels in edge detection. Following the metric proposed by the DoG filter, standard deviations $\sigma_1$ and $\sigma_2$ are setted, with $\sigma_2$ smaller than $\sigma_1$. 

\begin{figure}[!htbp]
	\center
	\includegraphics[width=.42\textwidth]{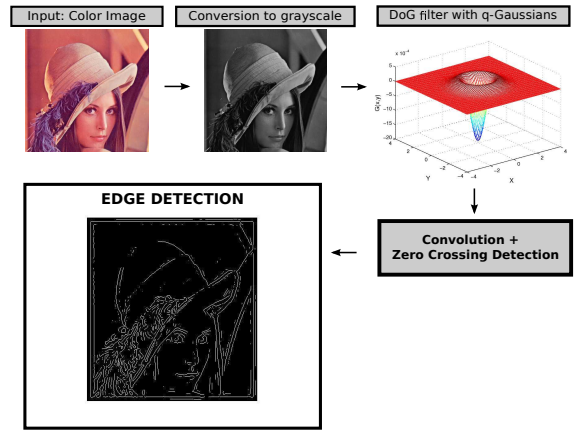}
	\caption{Algorithm for edge detection using the DoG method with q-Gaussian kernels.}
	\label{fig:7}
\end{figure}

After the filter having the appropriate size, we should set it with the input image in gray scale. After the convolution we identify the edges by using the ``zero cross'' detector. Figure \ref{fig:7} summarizes the described process.

\section{Results:  Gaussian vs q-Gaussian Edge Detection}
\begin{figure*}[!htbp]
	\center
		\subfigure[$q=-2.5$]{\label{fig:8}\includegraphics[width= .32\textwidth]{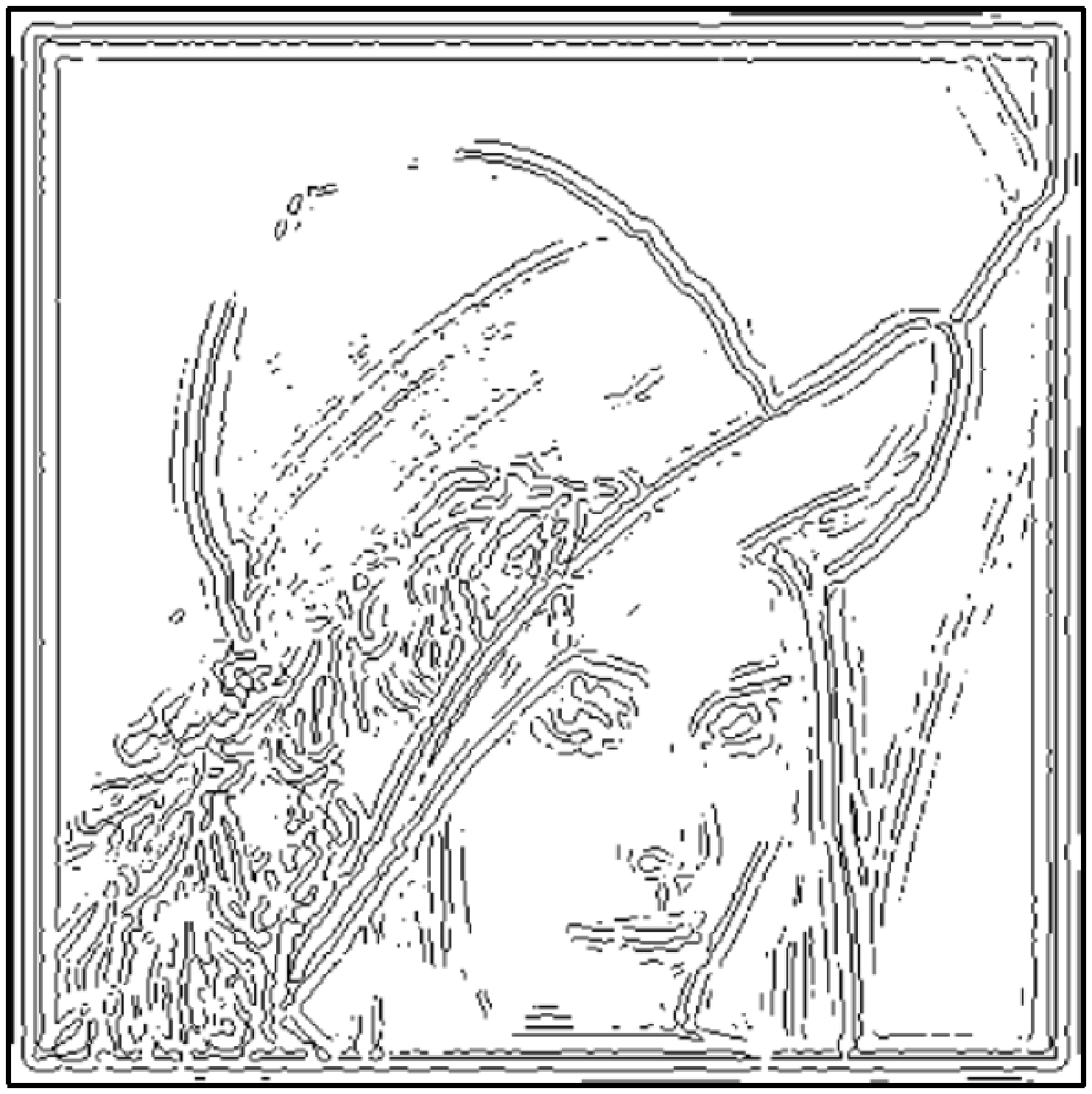}}
		\subfigure[$q=-1.625$]{\label{fig:8}\includegraphics[width= .32\textwidth]{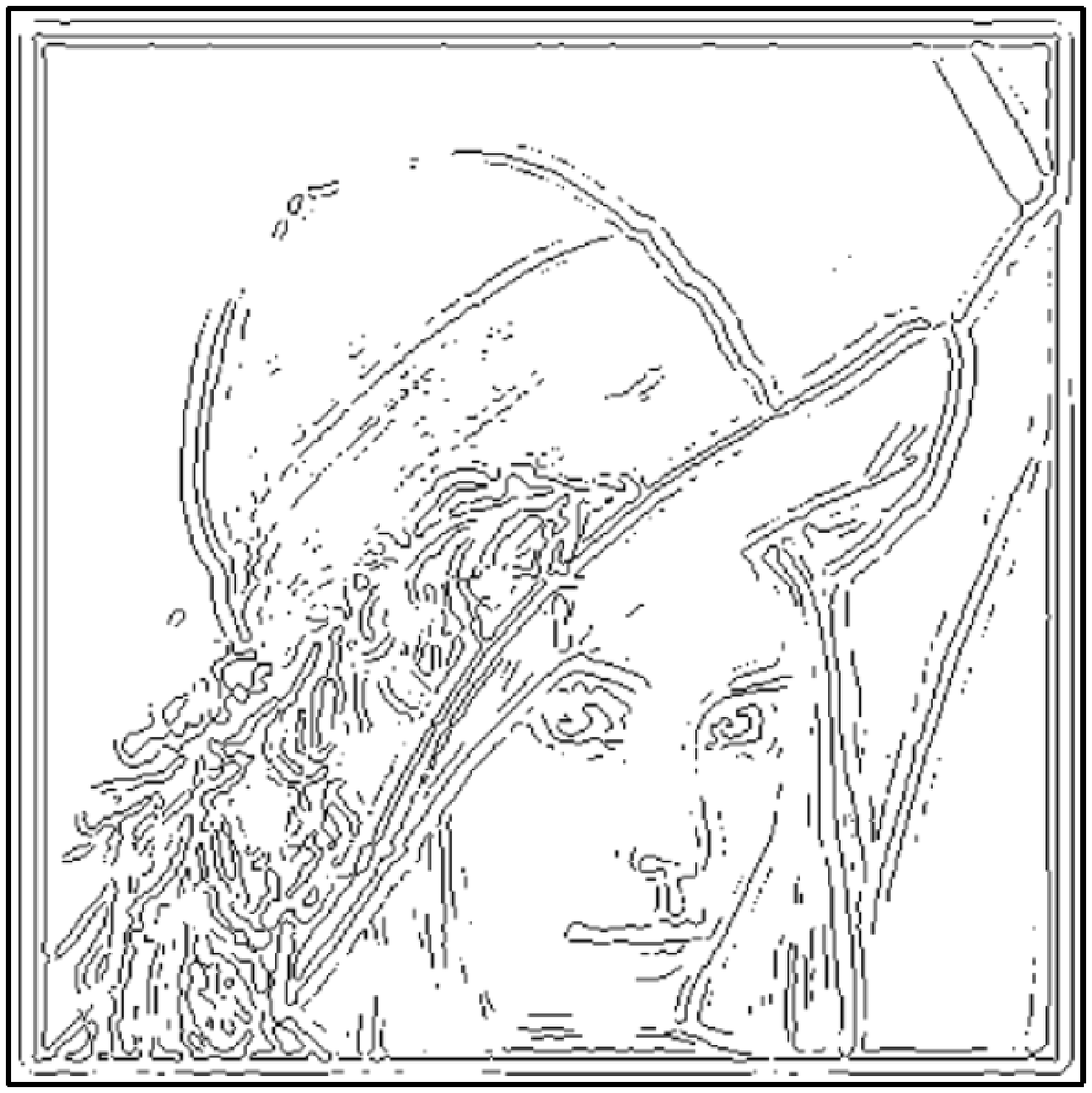}}
		\subfigure[$q=-0.75$]{\label{fig:8}\includegraphics[width= .32\textwidth]{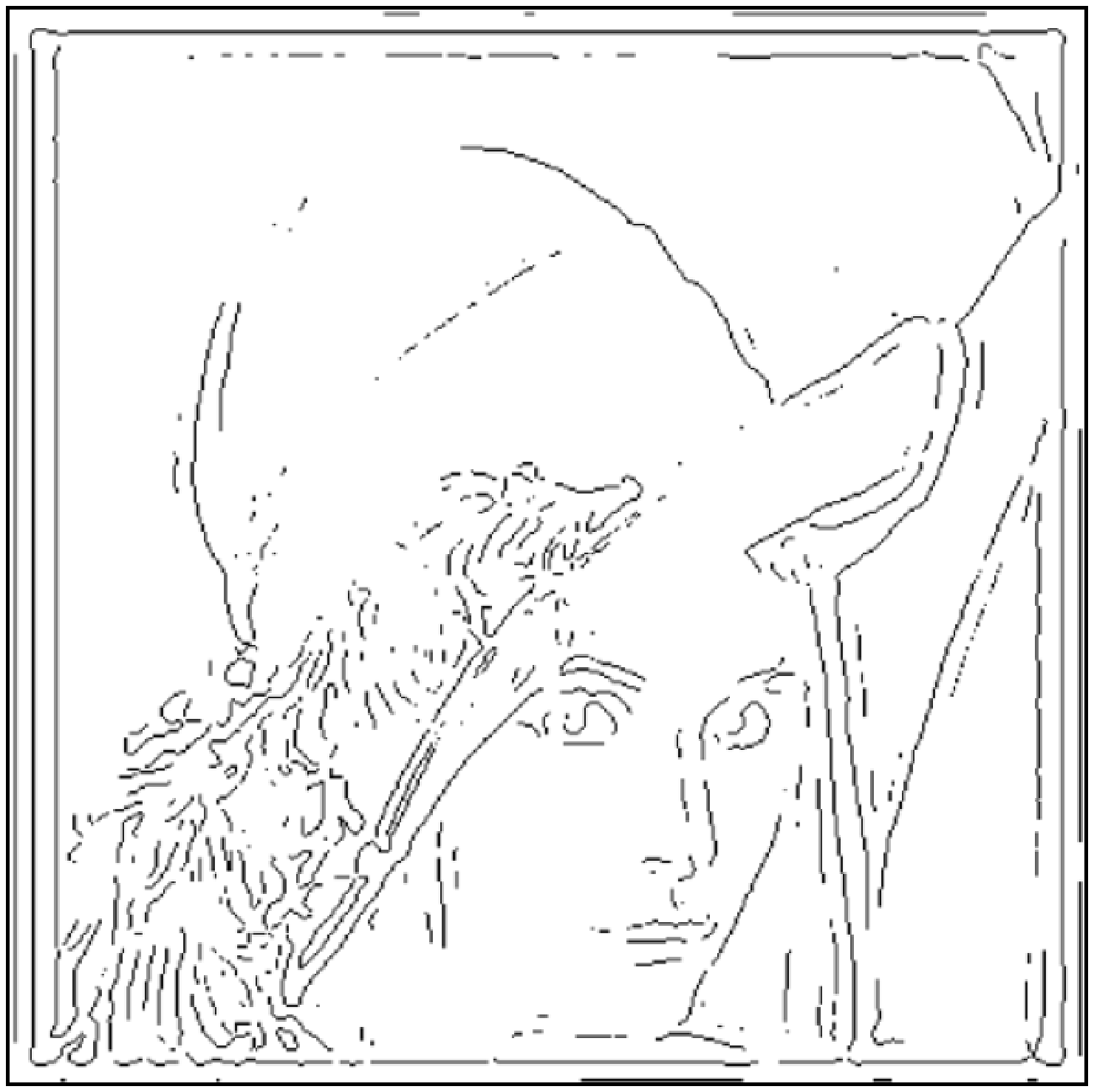}}
		\subfigure[$q=-0.125$]{\label{fig:8}\includegraphics[width= .32\textwidth]{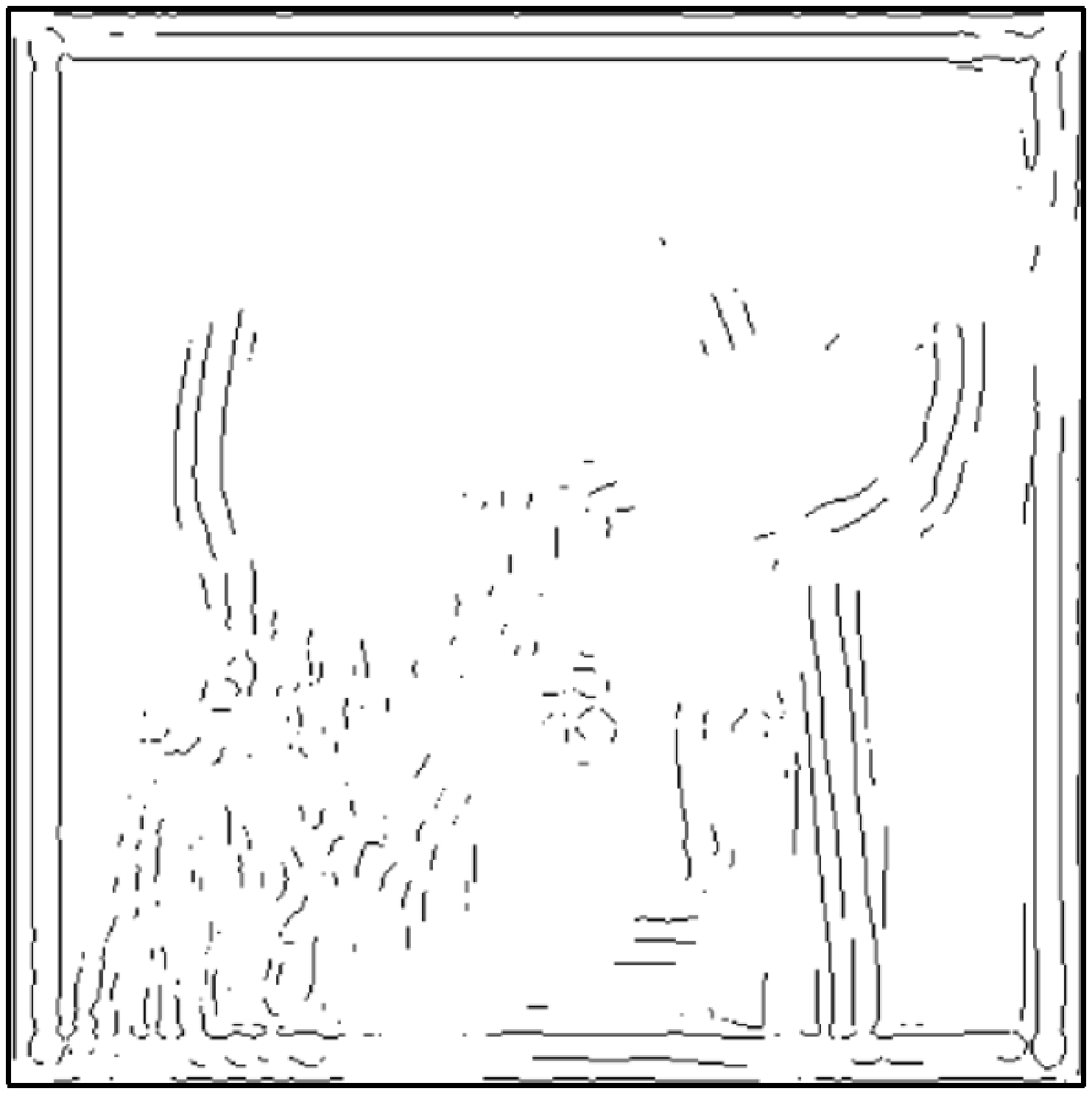}}
		\subfigure[$q=1$]{\label{fig:8}\includegraphics[width= .32\textwidth]{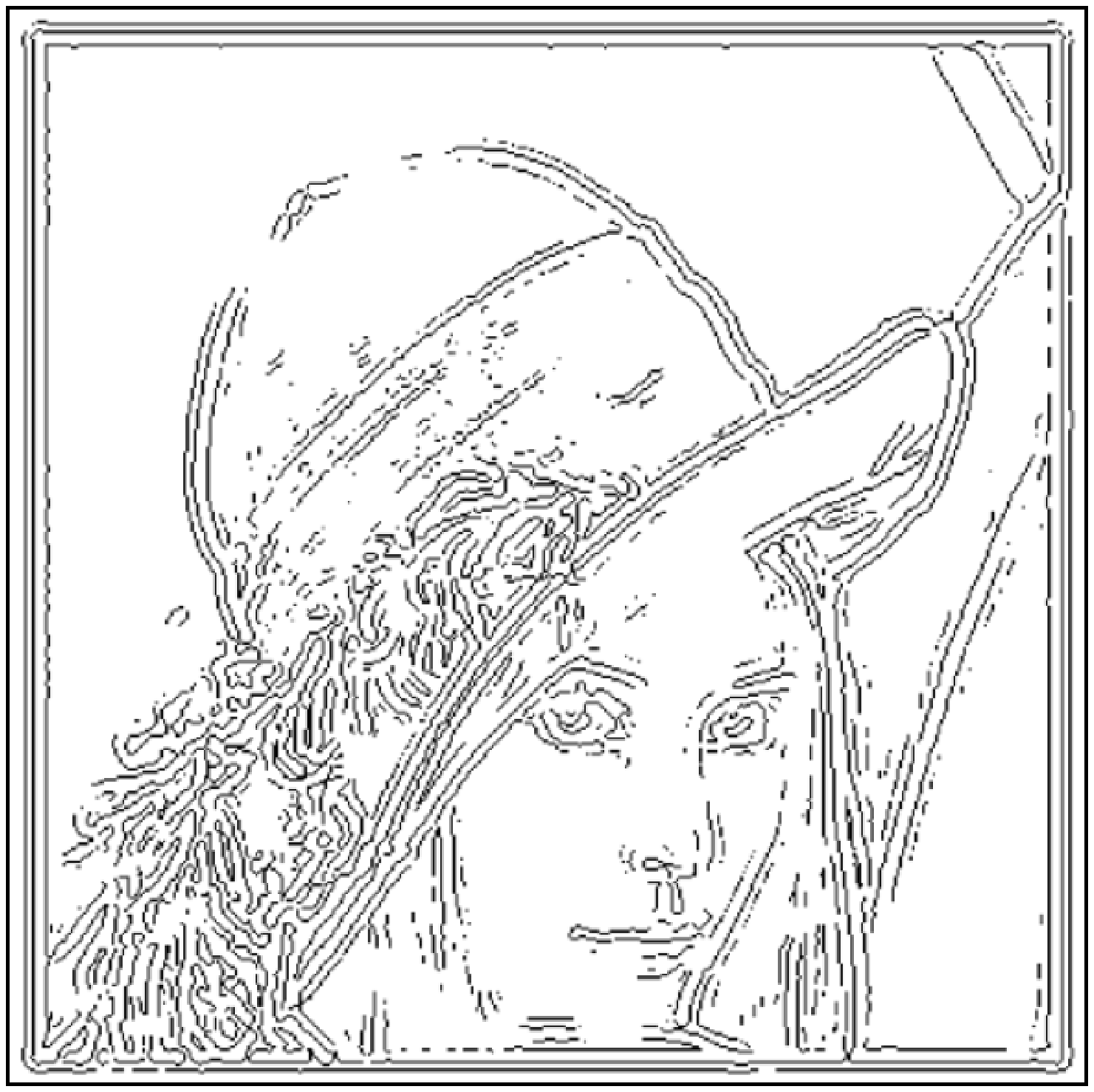}}
		\subfigure[$q=1.375$]{\label{fig:8}\includegraphics[width= .32\textwidth]{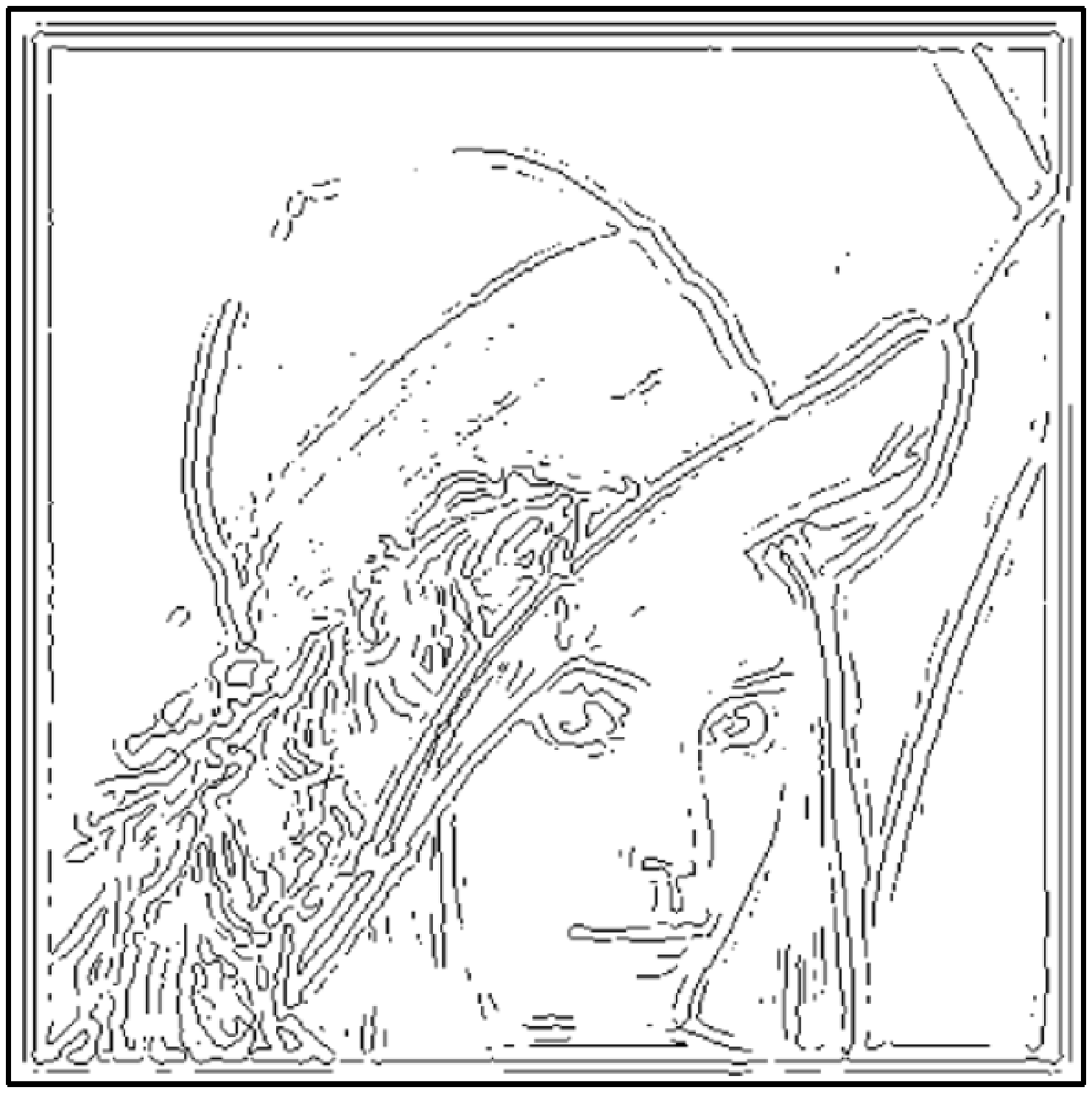}}
		\subfigure[$q=1.75$]{\label{fig:8}\includegraphics[width= .32\textwidth]{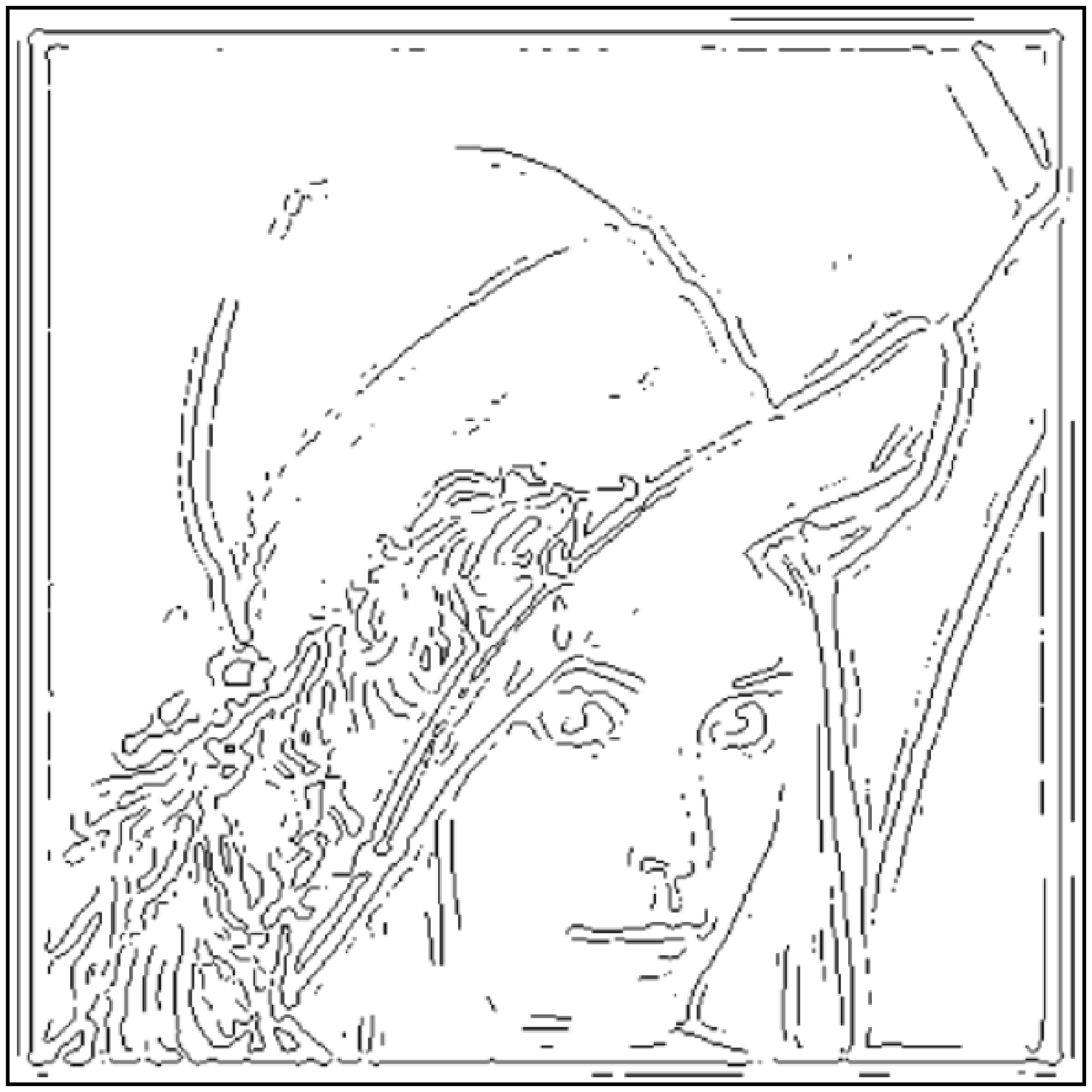}}
		\subfigure[$q=2.125$]{\label{fig:8}\includegraphics[width= .32\textwidth]{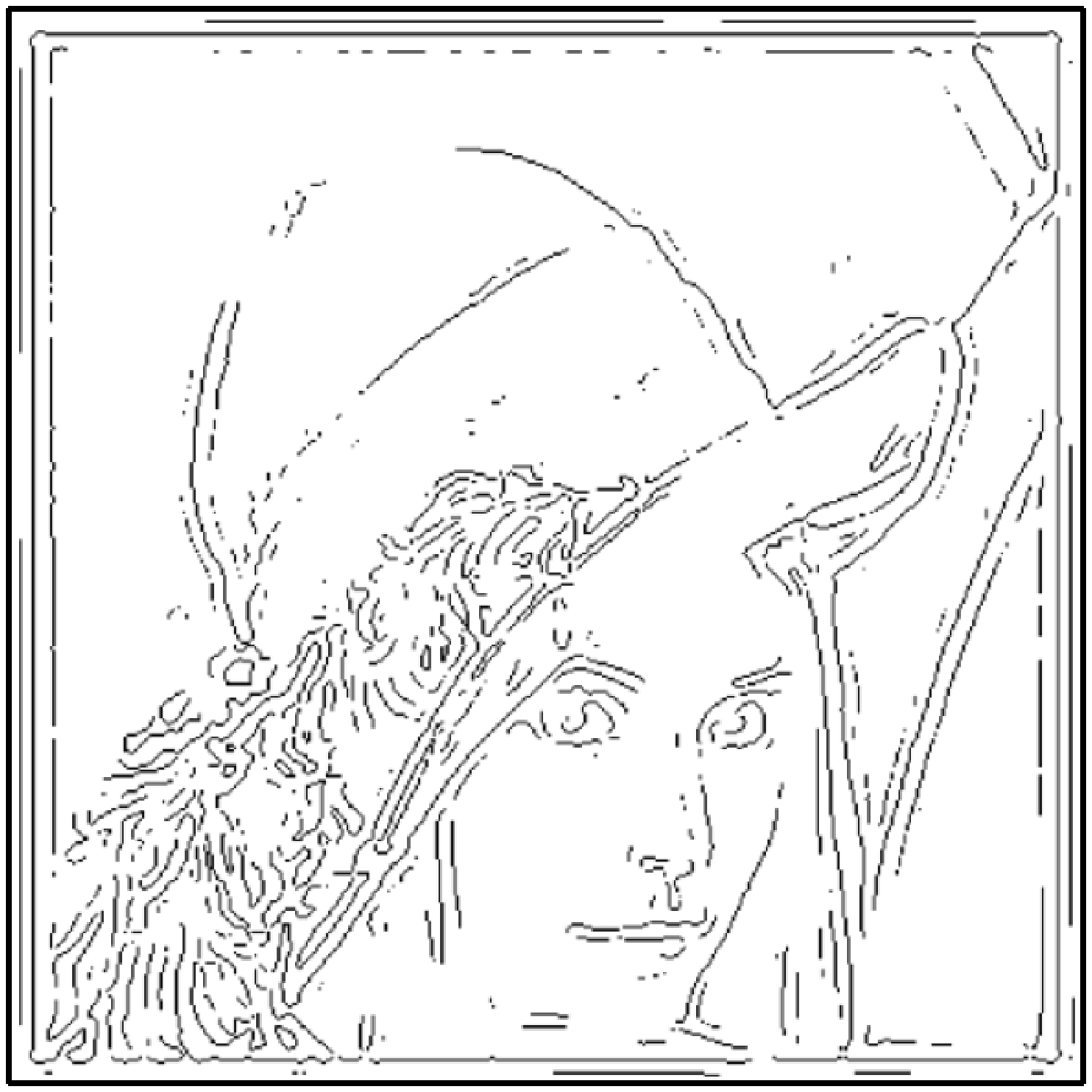}}
		\subfigure[$q=2.5$]{\label{fig:8}\includegraphics[width= .32\textwidth]{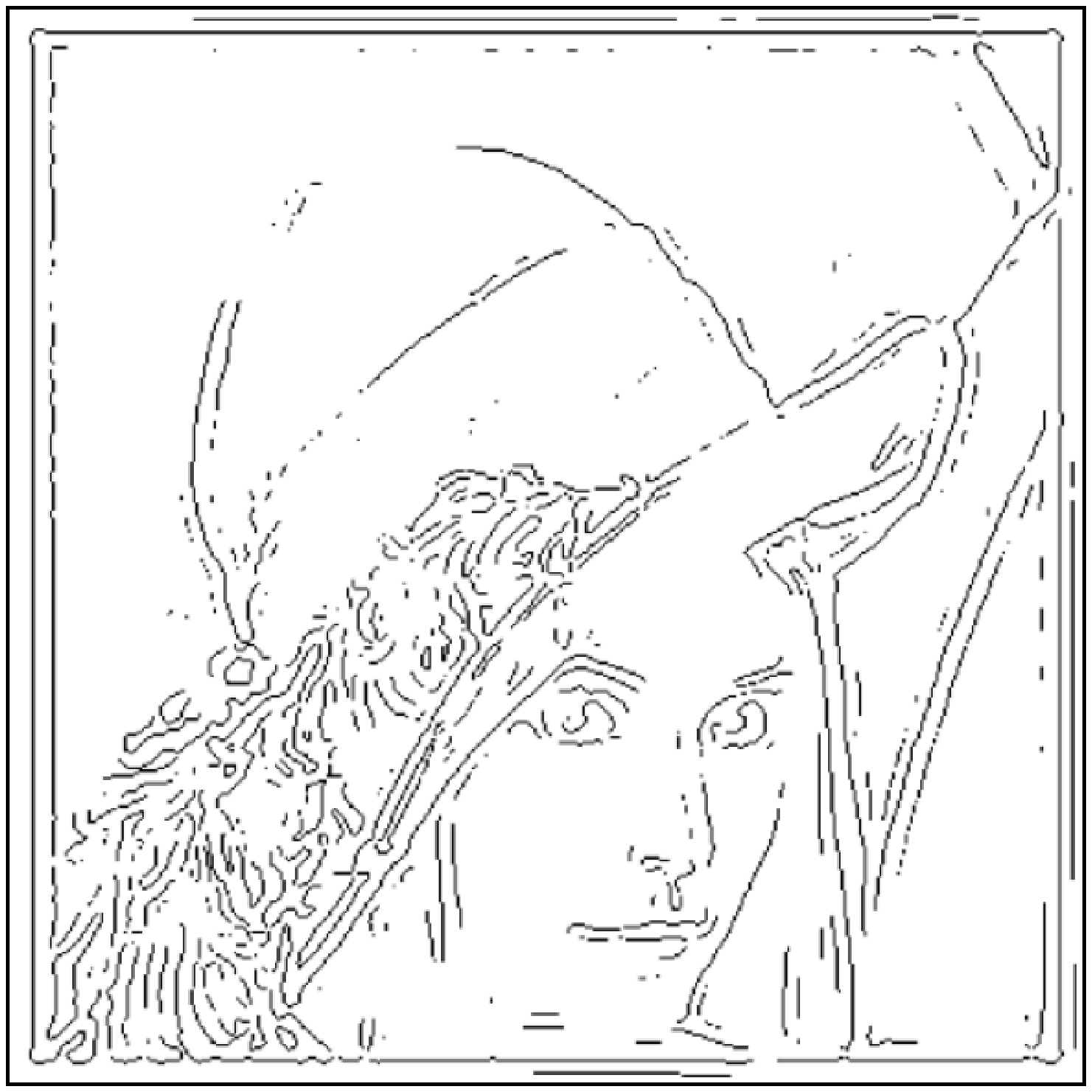}}
	\caption{DoG with different q-Gaussian kernels. Notice that for $q=1$, the q-Gaussian is exactly the standard one}
	\label{fig:Dog gaussians}
\end{figure*}

The results for edge detection using the method DoG with q-Gaussian kernels show up rich in detail when compared to the method DoG with Classic Gaussian kernels because the q-Gaussian probability distribution have the adjustable parameter $q$. This parameter allow us to define the degree of detail that we seek in our detection. Figure \ref{fig:Dog gaussians} shows results obtained from q-Gaussian using $\sigma_1 = 0.2$ and $\sigma_2 = 0.1$.

\section{Conclusions}
The results presented in this work show that using the DoG filter with q-Gaussian kernels proves to be an excellent alternative to the LoG and DoG with classical Gaussian kernels. Compared to the LoG filter, the proposed method has lower computational cost. Constructing the convolution mask from the subtraction of two Gaussian kernels with standard deviations $\sigma_1$ and $\sigma_2$ slightly different is much less costly than take the Laplacian of the q-Gaussian function (which involves calculating derivatives).

Compared to the DoG filter with kernels using the normal distribution of probabilities, we note that we gain in details of edge detection. That is because in addition to the variable parameter $\sigma$, responsible for more or less blurring (Gaussian blur), we also have the entropic index $q$, variable and responsible for the shape of q-Gaussian, being able to get more details that the traditional approach when both have the same blur.

The extensiveness or not extensiveness of the entropy depends on the system characteristics. Thus, it can be extended for certain values of $q$. In this point, we can apply this concept to our work. By using the q-Gaussian method, the entropic index $q$ allow us adjust the function used in the filter to get the details and results that are more relevant.

\newpage
\section*{Acknowledgments}
Lucas Assirati acknowledges the Confederation of Associations in the Private Employment Sector (CAPES) Grant . N\'ubia R. Silva, Lilian Berton and Odemir M. Bruno are grateful for S\~ao Paulo Research Foundation, grant Nos.: 2011/21467-9, 2011/21880-3 and 2011/23112-3. Bruno also acknowledges the National Council for Scientific and Technological Development (CNPq), grant Nos. 308449/2010-0 and 473893/2010-0.

\section*{References}

\begin{thebibliography}{13}%
\makeatletter
\providecommand \@ifxundefined [1]{%
 \@ifx{#1\undefined}
}%
\providecommand \@ifnum [1]{%
 \ifnum #1\expandafter \@firstoftwo
 \else \expandafter \@secondoftwo
 \fi
}%
\providecommand \@ifx [1]{%
 \ifx #1\expandafter \@firstoftwo
 \else \expandafter \@secondoftwo
 \fi
}%
\providecommand \natexlab [1]{#1}%
\providecommand \enquote  [1]{``#1''}%
\providecommand \bibnamefont  [1]{#1}%
\providecommand \bibfnamefont [1]{#1}%
\providecommand \citenamefont [1]{#1}%
\providecommand \href@noop [0]{\@secondoftwo}%
\providecommand \href [0]{\begingroup \@sanitize@url \@href}%
\providecommand \@href[1]{\@@startlink{#1}\@@href}%
\providecommand \@@href[1]{\endgroup#1\@@endlink}%
\providecommand \@sanitize@url [0]{\catcode `\\12\catcode `\$12\catcode
  `\&12\catcode `\#12\catcode `\^12\catcode `\_12\catcode `\%12\relax}%
\providecommand \@@startlink[1]{}%
\providecommand \@@endlink[0]{}%
\providecommand \url  [0]{\begingroup\@sanitize@url \@url }%
\providecommand \@url [1]{\endgroup\@href {#1}{\urlprefix }}%
\providecommand \urlprefix  [0]{URL }%
\providecommand \Eprint [0]{\href }%
\@ifxundefined \urlstyle {%
  \providecommand \doi  [0]{\begingroup \@sanitize@url \@doi}%
  \providecommand \@doi [1]{\endgroup \@@startlink {\doibase
  #1}doi:\discretionary {}{}{}#1\@@endlink }%
}{%
  \providecommand \doi  [0]{doi:\discretionary{}{}{}\begingroup
  \urlstyle{rm}\Url }%
}%
\providecommand \doibase [0]{http://dx.doi.org/}%
\providecommand \Doi [0]{\begingroup \@sanitize@url \@Doi }%
\providecommand \@Doi  [1]{\endgroup\@@startlink{\doibase#1}\@@Doi}%
\providecommand \@@Doi [1]{#1\@@endlink}%
\providecommand \selectlanguage [0]{\@gobble}%
\providecommand \bibinfo  [0]{\@secondoftwo}%
\providecommand \bibfield  [0]{\@secondoftwo}%
\providecommand \translation [1]{[#1]}%
\providecommand \BibitemOpen [0]{}%
\providecommand \bibitemStop [0]{}%
\providecommand \bibitemNoStop [0]{.\EOS\space}%
\providecommand \EOS [0]{\spacefactor3000\relax}%
\providecommand \BibitemShut  [1]{\csname bibitem#1\endcsname}%
\bibitem [{\citenamefont {Gudmundsson}\ \emph {et~al.}(1998)\citenamefont
  {Gudmundsson}, \citenamefont {El-Kwae},\ and\ \citenamefont
  {Kabuka}}]{Gudmundsson1998}%
  \BibitemOpen
  \bibfield  {author} {\bibinfo {author} {\bibfnamefont {M.}~\bibnamefont
  {Gudmundsson}}, \bibinfo {author} {\bibfnamefont {E.}~\bibnamefont
  {El-Kwae}}, \ and\ \bibinfo {author} {\bibfnamefont {M.}~\bibnamefont
  {Kabuka}},\ }\bibfield  {title} {\enquote {\bibinfo {title} {Edge detection
  in medical images using a genetic algorithm},}\ }\href@noop {} {\bibfield
  {journal} {\bibinfo  {journal} {Medical Imaging, IEEE Transactions on},\
  }\textbf {\bibinfo {volume} {17}},\ \bibinfo {pages} {469--474} (\bibinfo
  {year} {1998})}\BibitemShut {NoStop}%
\bibitem [{\citenamefont {Augusto}\ \emph {et~al.}(1984)\citenamefont
  {Augusto}, \citenamefont {Goltz},\ and\ \citenamefont
  {Dem\'{\i}sio}}]{Augusto1984}%
  \BibitemOpen
  \bibfield  {author} {\bibinfo {author} {\bibfnamefont {G.}~\bibnamefont
  {Augusto}}, \bibinfo {author} {\bibfnamefont {M.}~\bibnamefont {Goltz}}, \
  and\ \bibinfo {author} {\bibfnamefont {J.}~\bibnamefont {Dem\'{\i}sio}},\
  }\bibfield  {title} {\enquote {\bibinfo {title} {Detec\c{c}\~{a}o de bordas
  em imagens a\'{e}reas e de sat\'{e}lite com uso de redes neurais
  artificiais},}\ }\href@noop {} {,\ \bibinfo {pages} {1044--1045} (\bibinfo
  {year} {1984})}\BibitemShut {NoStop}%
\bibitem [{\citenamefont {Jain}\ \emph {et~al.}(1995)\citenamefont {Jain},
  \citenamefont {Kasturi},\ and\ \citenamefont {Schunck}}]{jain1995machine}%
  \BibitemOpen
  \bibfield  {author} {\bibinfo {author} {\bibfnamefont {R.}~\bibnamefont
  {Jain}}, \bibinfo {author} {\bibfnamefont {R.}~\bibnamefont {Kasturi}}, \
  and\ \bibinfo {author} {\bibfnamefont {B.}~\bibnamefont {Schunck}},\
  }\href@noop {} {\emph {\bibinfo {title} {Machine vision}}}\ (\bibinfo
  {publisher} {McGraw-Hill},\ \bibinfo {year} {1995})\BibitemShut {NoStop}%
\bibitem [{\citenamefont {Gonzalez}\ and\ \citenamefont
  {Woods}(2011)}]{gonzalez2011digital}%
  \BibitemOpen
  \bibfield  {author} {\bibinfo {author} {\bibfnamefont {R.}~\bibnamefont
  {Gonzalez}}\ and\ \bibinfo {author} {\bibfnamefont {R.}~\bibnamefont
  {Woods}},\ }\href@noop {} {\emph {\bibinfo {title} {Digital Image
  Processing}}}\ (\bibinfo  {publisher} {Pearson Education},\ \bibinfo {year}
  {2011})\BibitemShut {NoStop}%
\bibitem [{\citenamefont {de~Albuquerque}\ \emph {et~al.}(2004)\citenamefont
  {de~Albuquerque}, \citenamefont {Esquef},\ and\ \citenamefont
  {Mello}}]{PortesdeAlbuquerque20041059}%
  \BibitemOpen
  \bibfield  {author} {\bibinfo {author} {\bibfnamefont {M.~P.}\ \bibnamefont
  {de~Albuquerque}}, \bibinfo {author} {\bibfnamefont {I.~A.}\ \bibnamefont
  {Esquef}}, \ and\ \bibinfo {author} {\bibfnamefont {A.~R.~G.}\ \bibnamefont
  {Mello}},\ }\bibfield  {title} {\enquote {\bibinfo {title} {Image
  thresholding using tsallis entropy},}\ }\href@noop {} {\bibfield  {journal}
  {\bibinfo  {journal} {Pattern Recognition Letters},\ }\textbf {\bibinfo
  {volume} {25}},\ \bibinfo {pages} {1059 -- 1065} (\bibinfo {year}
  {2004})}\BibitemShut {NoStop}%
\bibitem [{\citenamefont {Sathya}\ and\ \citenamefont
  {Kayalvizhi}(2010)}]{Sathya2010}%
  \BibitemOpen
  \bibfield  {author} {\bibinfo {author} {\bibfnamefont {P.~D.}\ \bibnamefont
  {Sathya}}\ and\ \bibinfo {author} {\bibfnamefont {R.}~\bibnamefont
  {Kayalvizhi}},\ }\bibfield  {title} {\enquote {\bibinfo {title} {Pso-based
  tsallis thresholding selection procedure for image segmentation},}\
  }\href@noop {} {\bibfield  {journal} {\bibinfo  {journal} {Pattern
  Recognition Letters},\ }\textbf {\bibinfo {volume} {5}},\ \bibinfo {pages}
  {39 -- 46} (\bibinfo {year} {2010})}\BibitemShut {NoStop}%
\bibitem [{\citenamefont {Kilic}\ and\ \citenamefont
  {Kayacan}(2012)}]{Kilic2012}%
  \BibitemOpen
  \bibfield  {author} {\bibinfo {author} {\bibfnamefont {I.}~\bibnamefont
  {Kilic}}\ and\ \bibinfo {author} {\bibfnamefont {O.}~\bibnamefont
  {Kayacan}},\ }\bibfield  {title} {\enquote {\bibinfo {title} {Generalized
  \{ICM\} for image segmentation based on tsallis statistics},}\ }\href@noop {}
  {\bibfield  {journal} {\bibinfo  {journal} {Physica A: Statistical Mechanics
  and its Applications},\ }\textbf {\bibinfo {volume} {391}},\ \bibinfo {pages}
  {4899 -- 4908} (\bibinfo {year} {2012})}\BibitemShut {NoStop}%
\bibitem [{\citenamefont {Fabbri}\ \emph {et~al.}(2013)\citenamefont {Fabbri},
  \citenamefont {Bastos}, \citenamefont {Neto}, \citenamefont {Lopes},\ and\
  \citenamefont {Bruno}}]{Fabbri2013}%
  \BibitemOpen
  \bibfield  {author} {\bibinfo {author} {\bibfnamefont {R.}~\bibnamefont
  {Fabbri}}, \bibinfo {author} {\bibfnamefont {I.}~\bibnamefont {Bastos}},
  \bibinfo {author} {\bibfnamefont {F.}~\bibnamefont {Neto}}, \bibinfo {author}
  {\bibfnamefont {F.}~\bibnamefont {Lopes}}, \ and\ \bibinfo {author}
  {\bibfnamefont {O.}~\bibnamefont {Bruno}},\ }\bibfield  {title} {\enquote
  {\bibinfo {title} {Multi-q pattern classification of polarization curves},}\
  }\href@noop {} { (\bibinfo {year} {2013})},\ \Eprint
  {http://arxiv.org/abs/arXiv:1305.2876 [cs.CE]} {arXiv:arXiv:1305.2876
  [cs.CE]} \BibitemShut {NoStop}%
\bibitem [{\citenamefont {Fabbri}\ \emph {et~al.}(2012)\citenamefont {Fabbri},
  \citenamefont {Gon\c{c}alves}, \citenamefont {Lopes},\ and\ \citenamefont
  {Bruno}}]{Fabbri2012}%
  \BibitemOpen
  \bibfield  {author} {\bibinfo {author} {\bibfnamefont {R.}~\bibnamefont
  {Fabbri}}, \bibinfo {author} {\bibfnamefont {W.~N.}\ \bibnamefont
  {Gon\c{c}alves}}, \bibinfo {author} {\bibfnamefont {F.~J.}\ \bibnamefont
  {Lopes}}, \ and\ \bibinfo {author} {\bibfnamefont {O.~M.}\ \bibnamefont
  {Bruno}},\ }\bibfield  {title} {\enquote {\bibinfo {title} {Multi-q pattern
  analysis: A case study in image classification},}\ }\href@noop {} {\bibfield
  {journal} {\bibinfo  {journal} {Physica A: Statistical Mechanics and its
  Applications},\ }\textbf {\bibinfo {volume} {391}},\ \bibinfo {pages} {4487
  -- 4496} (\bibinfo {year} {2012})}\BibitemShut {NoStop}%
\bibitem [{\citenamefont {Soares}\ and\ \citenamefont
  {Murta}(2013)}]{Soares2013}%
  \BibitemOpen
  \bibfield  {author} {\bibinfo {author} {\bibfnamefont {I.~J.~A.}\
  \bibnamefont {Soares}}\ and\ \bibinfo {author} {\bibfnamefont {L.~O.}\
  \bibnamefont {Murta}},\ }\bibfield  {title} {\enquote {\bibinfo {title}
  {Noise reduction using nonadditive q-gaussian filters in magnetic resonance
  images},}\ }in\ \Doi {doi:10.1117/12.2007119} {\emph {\bibinfo {booktitle}
  {Proc. SPIE 8669, Medical Imaging 2013: Image Processing, 86692J}}}\
  (\bibinfo {year} {2013})\BibitemShut {NoStop}%
\bibitem [{\citenamefont {Tsallis}(1988)}]{Tsallis:1988ws}%
  \BibitemOpen
  \bibfield  {author} {\bibinfo {author} {\bibfnamefont {C.}~\bibnamefont
  {Tsallis}},\ }\bibfield  {title} {\enquote {\bibinfo {title} {Possible
  generalization of boltzmann-gibbs statistics},}\ }\href@noop {} {\bibfield
  {journal} {\bibinfo  {journal} {Journal of Statistical Physics},\ }\textbf
  {\bibinfo {volume} {52}},\ \bibinfo {pages} {479--487} (\bibinfo {year}
  {1988})}\BibitemShut {NoStop}%
\bibitem [{\citenamefont {Shannon}(1948)}]{Shannon:1948wk}%
  \BibitemOpen
  \bibfield  {author} {\bibinfo {author} {\bibfnamefont {C.~E.}\ \bibnamefont
  {Shannon}},\ }\bibfield  {title} {\enquote {\bibinfo {title} {A mathematical
  theory of communication},}\ }\href@noop {} {\bibfield  {journal} {\bibinfo
  {journal} {The Bell System Technical Journal},\ }\textbf {\bibinfo {volume}
  {27}},\ \bibinfo {pages} {379--423--623--656} (\bibinfo {year}
  {1948})}\BibitemShut {NoStop}%
\bibitem [{\citenamefont {Tsallis}(2011)}]{Tsallis:2011hr}%
  \BibitemOpen
  \bibfield  {author} {\bibinfo {author} {\bibfnamefont {C.}~\bibnamefont
  {Tsallis}},\ }\bibfield  {title} {\enquote {\bibinfo {title} {The nonadditive
  entropy sq and its applications in physics and elsewhere: Some remarks},}\
  }\href@noop {} {\bibfield  {journal} {\bibinfo  {journal} {Entropy},\
  }\textbf {\bibinfo {volume} {13}},\ \bibinfo {pages} {1765--1804} (\bibinfo
  {year} {2011})}\BibitemShut {NoStop}%
\end{thebibliography}

%


\end{document}